\documentclass{scrartcl}

\usepackage{geometry}
\usepackage{algorithm}
\usepackage{algpseudocode}
\usepackage{bbm}

\usepackage{blindtext}
\usepackage[T1]{fontenc}
\usepackage[utf8]{inputenc}

\usepackage{amsmath, amsfonts, amsthm, amssymb}
\usepackage{braket, nicefrac}

\usepackage{siunitx}

\usepackage{enumitem, multicol}

\usepackage{graphicx, float}  
\usepackage{pgfplots}\usepgfplotslibrary{units}\pgfplotsset{compat=1.16}

\usepackage{setspace}

\usepackage{hyperref}

\graphicspath{{graphics/}{Graphics/}{./}}

\renewenvironment{abstract}{
    \begin{center}
    \textbf{Abstract}
    \vspace{0.5cm}
    \par\itshape
    \begin{minipage}{0.8\linewidth}}{\end{minipage}
    \noindent\ignorespaces
    \end{center}
}

\newenvironment{keywords}{
    \begin{center}
    \textbf{Keywords}
    \vspace{0.5cm}
    \par
    \begin{minipage}{0.8\linewidth}}{\end{minipage}
    \noindent\ignorespaces
    \end{center}
}

\newenvironment{preface}{
    \begin{center}
    \textbf{Preface}
    \vspace{0.5cm}
    \par
    \begin{minipage}{0.8\linewidth}}{\end{minipage}
    \noindent\ignorespaces
    \end{center}
}

\newenvironment{acknowledgements}{
    \begin{center}
    \textbf{Acknowledgements}
    \vspace{0.5cm}
    \par
    \begin{minipage}{0.8\linewidth}}{\end{minipage}
    \noindent\ignorespaces
    \end{center}
}

\begin{document}
\begin{titlepage}
	\centering
	\includegraphics[width=0.6\textwidth]{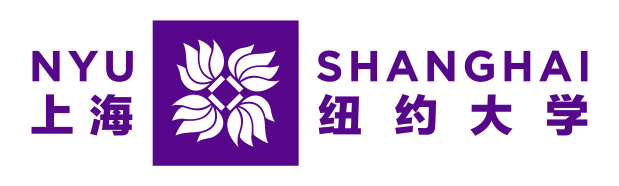}\par
	\vspace{2cm}
	{\scshape\LARGE Computer Science \par}  
	{\scshape\LARGE \& \par}                
	{\scshape\LARGE Data Science \par}      
	\vspace{1cm}
	{\scshape\Large Capstone Report - Fall 2022\par}
	\vfill
	
	{\huge\bfseries Controllable Ancient Chinese Lyrics Generation Based on Phrase Prototype Retrieving\par}
	\vfill
	
	{\Large\itshape Li Yi\\}\par
	\vspace{1.5cm}

	\vfill
	supervised by\par
	Wilson Tam

	\vfill
\end{titlepage}

\newpage

\begin{preface}
        This report proposes a novel method for generating ancient Chinese lyrics (Song Ci), a type of ancient lyrics that involves precise control of song structure. The method is equipped with a phrase retriever and a phrase connector that select and concatenates phrases. Additionally, this method should be able to generalize in other lyrics generation tasks.
\end{preface}

\vspace{1cm}

\begin{acknowledgements}
Throughout the research process, I have received great support and assistance. I would first like to thank my supervisor, Prof. Wilson Tam, whose expertise was invaluable in formulating the research questions and methodology. Your insightful feedback pushed me to sharpen my thinking and brought my work to a higher level. I would also like to acknowledge Prof. Gus Xia, who provided me with the initial idea. Thanks to all the participants in the subject evaluation of this research. Thanks to my friends and parents, who have supported me throughout the process.

\end{acknowledgements}

\newpage

\begin{abstract}
Generating lyrics and poems is one of the essential downstream tasks in the Natural Language Processing (NLP) field. Current methods have performed well in some lyrics generation scenarios but need further improvements in tasks requiring fine control. We propose a novel method for generating ancient Chinese lyrics (Song Ci), a type of ancient lyrics that involves precise control of song structure. The system is equipped with a phrase retriever and a phrase connector. Based on an input prompt, the phrase retriever picks phrases from a database to construct a phrase pool. The phrase connector then selects a series of phrases from the phrase pool that minimizes a multi-term loss function that considers rhyme, song structure, and fluency. Experimental results show that our method can generate high-quality ancient Chinese lyrics while performing well on topic and song structure control. We also expect our approach to be generalized to other lyrics-generating tasks.
\end{abstract}
\vspace{1cm}


\newpage

\doublespacing
\tableofcontents
\singlespacing

\newpage

\doublespacing

\section{Introduction}

Recently, the field of Natural Language Processing (NLP) has significantly developed because of new emerging deep learning models and architectures. Transformer-based large-scale pre-trained models, including BERT \cite{bert_devlin_chang_lee_toutanova} and GPT \cite{gpt3_2020}, have achieved fantastic results in multiple downstream tasks. Lyrics generation is one of the essential downstream tasks, and previous research has utilized large-scale models in this task and has achieved good results \cite{ram_gummadi_bhethanabotla_savery_weinberg_2021,zhang_lapata}. Compared with typical language generation tasks, lyrics generation usually involves the control of rhyme and sometimes the number of words in a sentence. Hence, current models still have some limitations: 1) The controllability is still weak. Many current models generate only based on the first sentence/phrase and do not involve the control of the number of words in each sentence or the rhyme. 2) Lyrics easily get unclear and off-topic after a few sentences.

In this project, we aim to develop a new system that can generate meaningful lyrics and achieve high controllability simultaneously. In particular, we limit the target of the generation to Ancient Chinese Lyrics (Song Ci), a type of ancient poem that involves precise control of song structure. However, we expect our approach to be generalized to other lyrics-generating tasks. The system is designed to generate meaningful lyrics that fit in a given prompt which contains both the song topic and the structure. Inspired by the fact that poems and lyrics are usually created by piling up literary images, we equip the standard large-scale language models with a phrase retriever and a phrase connector. Based on an input prompt, the phrase retriever picks phrases from a self-created phrase database to construct a phrase candidate pool. The phrase connector then selects a series of phrases from the phrase pool that minimizes a multi-term loss function that considers rhyme, song structure, and fluency. Experimental results show that our method can generate high-quality ancient Chinese lyrics while performing well on topic and song structure control.

In sum, the contributions of our research are as follows:

\begin{enumerate}
    \item A novel phrase-retrieving based method for ancient Chinese lyrics and general lyrics generation.
    \item A phrase connecting algorithm that can generate high-quality and controllable lyrics from a phrase pool.
    \item A labeled ancient Chinese lyrics phrase dataset that can be used for semantic analysis and ancient text generation.
\end{enumerate}

\section{Related Work}

\subsection{General Text Generation}
Generating lyrics with machines has always been an interesting topic that intersects natural language processing, musicology, and literary arts. Generally speaking, lyrics generation is one of the specific tasks regarding text generation, where multiple language models have been evaluated. Recently, deep neural network-based models have been proposed and achieve good results, including Recurrent Neural Networks (RNN)\cite{rnn_jozefowicz_2016}, Long Short-Term Memory (LSTM) \cite{textgan_zhang_2017}, and General Adversarial Networks (GAN) \cite{gan_guo_2017}. After the emergence of Transformers \cite{attention_vaswani_2017} that solves the memory issue when generating longer text, previous works have developed multiple methods based on this architecture, including auto-regressive methods like GPT \cite{gpt3_2020}, and auto-encoder methods like BERT \cite{bert_devlin_chang_lee_toutanova}. These methods greatly improve the performance on long text generation and generalize well to other downstream tasks. However, auto-regressive models cannot utilize the entire context when generating and also achieve low controllability. Previous research has adopted a continuous diffusion-based approach \cite{diffusion_li_2022} and improves the performance on challenging fine-grained control tasks.

\subsection{Ancient Lyrics Generation}\label{AncientLyricsGeneration}
Ancient Chinese lyrics (\textit{Ci} in Chinese) is a special type of literal art in ancient China, and only a few research especially focuses on generating ancient Chinese lyrics. However, broadly speaking, ancient Chinese poems and ancient lyrics generation in other languages are similar to the current task and are regarded as relevant fields. The research on ancient lyrics/poem generation started very early, even before the large-scale utilization of neural networks. Early-stage approaches are mainly template/phrase-based. One of the commonly used methods is simply searching for phrases and association words in a database according to some user inputs and concatenating them to produce the final lyrics/poem \cite{template_wu_tosa_nakatsu_2009}. Later on, some statistical machine learning algorithms are equipped based on the initial phrase selection approach, including Statistical Machine Translation (SMT) \cite{smt_2021}, and generative summarization \cite{summarization_2013}. The above methods have generated good results, but most of them are still weak in achieving full-piece coherency and fluency. 

Recently, deep learning models and architectures have proved to be promising in multiple fields, previous research also adopts the idea into ancient lyrics generation. One approach is still using phrase selection to formulate the first sentence, then generate the subsequent sentence according to the history sentences by RNNs auto-regressively \cite{rnn_jozefowicz_2016}. After that, more deep learning models were evaluated to further improve the fluency and poetic, including the planning-based neural network \cite{planning_wang_2016}, iterative polishing \cite{polishing_deng_2019}, and the conditional variational auto-encoders (CVAE) model \cite{cvae_yang_lin_suo_li_2020}. The above methods have achieved outstanding experimental results, but still, only a few are focusing on ancient lyrics generation, which is a harder task because it involves more structural constraints. At the same time, topic control is still a hard issue. In this paper, we try to combine the traditional methods and deep learning models and propose a method that contain mainly two stages: phrase retrieval and rewriting. The related works of which will be introduced in the next two subsections.

\subsection{Semantic Analysis and Phrase Retrieval}

In the first stage of our method, we want to retrieve phrases that are semantically related to the input prompt. One commonly used method is by computing the embedding of the characters, previous research have proposed word2vec \cite{word2vec_mikolov_2013} and GloVe \cite{glove_pennington_2014}, and also Sentence-BERT \cite{sentence_bert_reimers_gurevych_2019} that can calculate the embeddings sentence-wise. We adopt the above methods in this research.

\subsection{Utterance Editing and Rewriting}
In the second stage, we want to edit and rewrite the phrases to improve general fluency, rhyme consistency, and structural fitness. Previous research have proposed the Hierarchical Context Tagger (HTC) that solves the limitation of low convergence when add phrases \cite{utterance_rewriting_jin_2022}, and the context-aware neural editor \cite{prototype_editing_wu_2019} that applies Seq2Seq \cite{seq2seq} and the Attention mechanism \cite{attention_vaswani_2017}. These methods are used as references in this research.

\section{Solution}

\begin{figure}[H]
	\begin{center}
		\includegraphics[scale=0.15]{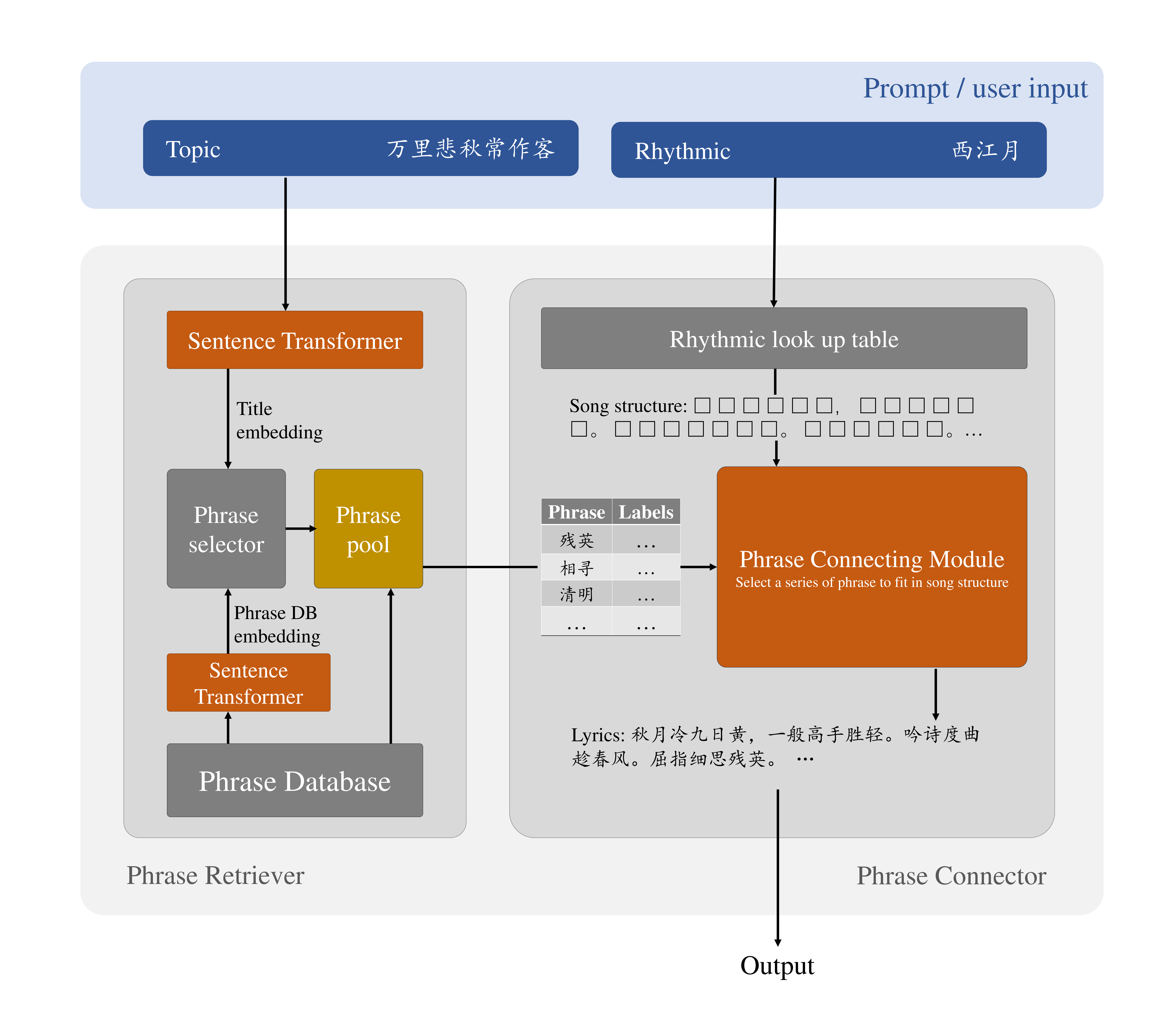}
	\end{center}
	\caption{The general architecture of the lyrics generating system}
	\label{fig:architec}
\end{figure}

The system diagram of our lyrics generating system is shown in Figure~\ref{fig:architec}. It can generate controllable ancient Chinese lyrics (Song Ci) according to a user input prompt. The prompt consist of two parts: the topic and the rhythmic. The topic specifies the main content the generated lyrics will be based on. The rhythmic, one of the unique attributes of ancient Chinese lyrics, is a conventional phrase that specifies the structure of a song (i.e., how many sentences, the number of characters in each sentence, the punctuation, etc.). When generating, the phrase retriever first picks phrases from a self-created phrase database according to the input topic. The phrase connector then selects a series of phrases from the constructed phrase pool that minimizes a multi-term loss function that considers rhyme, song structure, and fluency. For the rest of this section, we first introduce the self-created phrase dataset in Section \ref{DatasetCuration}. Then we describe the architecture of the phrase retriever in Section \ref{PhraseRetriever}. Finally, we show how the phrase connector is designed in Section \ref{PhraseConnector}.

\subsection{Dataset Curation}\label{DatasetCuration}

We create our dataset from an existing ancient Chinese lyrics dataset \textit{Quan Song Ci} (Complete ancient Chinese lyrics)\cite{ci}. The original dataset contains more than 21050 records of lyrics paragraphs, each labeled with a rhythmic and the author. Most records do not have a title, a common phenomenon for ancient Chinese lyrics. To construct our phrase database, we must split each sentence into phrases. However, how to determine the granularity is a problem. Small granularity will cause the system to degenerate into a character-level language model, while large granularity will result in low variance and creativity. 

In the actual implementation, we first construct the semantic tree of each sentence using HanLP\cite{hanlp}, an NLP toolkit that performs well on ancient Chinese. The sentences are split into morphemes while ensuring semantic accuracy. At this stage, the semantic tree contains many single-character and double-character phrases, which is different from our expectation. We further perform concatenation on the semantic tree. We traverse through the semantic tree using an in-order traversal. For each node, if the total phrase length of all its children is smaller than some threshold, we would combine the morphemes of all its children to make a single phrase. The algorithm is shown in Algorithm~\ref{alg::extractphrases}. The final created phrase dataset contains 514636 records, with an average phrase length of 2.75 characters. Each phrase is labeled with its original rhythmic, the source sentence position in the song, whether it is the beginning or the ending phrase of a sentence, etc.

\begin{algorithm}[h]
  \caption{Extract Phrases from Semantic Tree} 
  \label{alg::extractphrases}
  \begin{algorithmic}[1]
    \Require
      $n$: node structure of the semantic tree. Each node has a list of children and its content;
      $n_r$: root node of the semantic tree.
    \Ensure
      phrase list $p$
    \Function {ExtractPhrase}{n}
    \If {$n$ has no children} 
        \State $p[len(p)] \gets n.content$
        \State \Return
    \EndIf
    \For{$i=0 \to len(n.children)$} 
        \If {$n.children[i]$ has no children}
            \State $p[len(p)] \gets n.chilren[i].content$
        \ElsIf  {$len(\Call{ConcatNode}{n.children[i]}) <= 4$}
            \State $p[len(p)] \gets \Call{ConcatNode}{n.children[i]}$ 
        \Else 
            \State $\Call{ExtractPhrase}{n.children[i]}$
        \EndIf
    \EndFor
    \EndFunction
    \Function {ConcatNode}{$n$}
    \If {$n$ has no children}
        \Return {$n.content$}
    \Else
        \State $string \gets ""$
        \For{$i=0 \to len(n.children)$} $string \gets string + \Call{ConcatNode}{n.chilren[i]}$ 
        \EndFor
        \State \Return{TL}
    \EndIf
    \EndFunction
    \State $\Call{ExtractPhrase}{n_r}$
  \end{algorithmic}
\end{algorithm}

\subsection{Phrase Retriever}\label{PhraseRetriever}

We want to retrieve phrases that are semantically similar to the input topic. The Sentence Transformer model \cite{sentence_bert_reimers_gurevych_2019} can calculate sentence-wise semantics and convert each sentence to an embedding with 512 dimensions, which fits in our case. In the following sections, we denote a sentence using $s_i$, and a phrase whose source sentence is $s_i$ is denoted as $p_{ij}$, where $j$ specifies the phrase index in the sentence. To calculate semantics, we regard the source sentence's embedding as each phrase's embedding. That is to say, 
\begin{equation}
    Emb(p_{ij_m}) = Emb(p_{ij_n}) = Emb(s_i)
\end{equation}
where $j_m$ and $j_n$ are arbitrary phrase indexes in sentence $s_i$. Instead of calculating phrase-wise semantics, the above approach has the following advantages. 1) Extend the phrase candidate pool and avoid retrieving many phrases with exactly the same semantics as the topic. Failing to achieve this will result in the generated lyrics constantly repeating the same word. 2) Connectives will be contained in the constructed candidate pool. 

\begin{figure}[H]
	\begin{center}
		\includegraphics[scale=0.075]{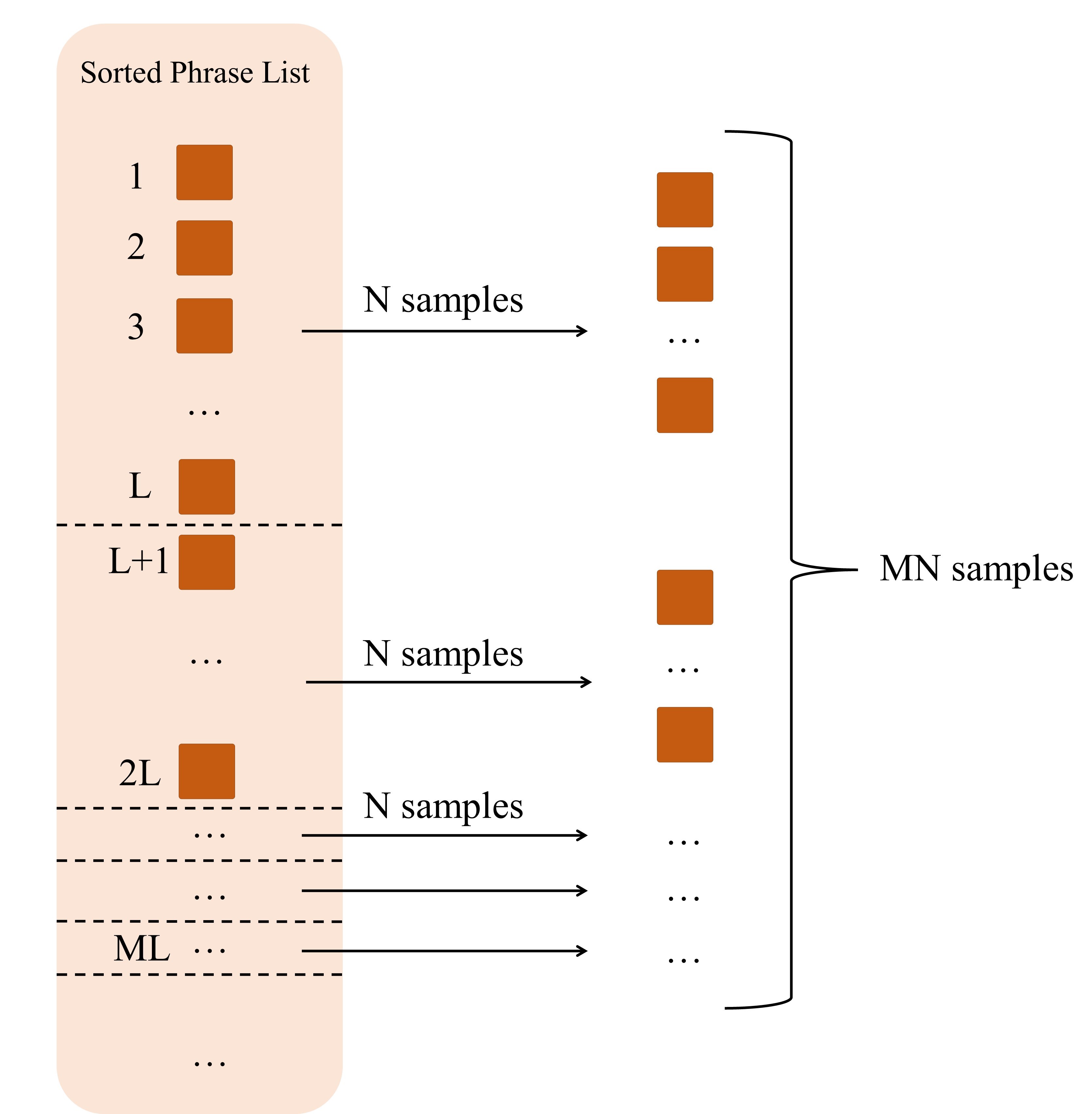}
	\end{center}
	\caption{The phrase selector select sample phrases from the sorted phrase list}
	\label{fig:retriever}
\end{figure}

We store the calculated embedding to speed up the retrieving time. When retrieving, the phrase selector would first sort the whole phrase database according to the cosine similarity between the topic embedding and each phrase embedding in descending order. Then, starting from the most similar phrase, the phrase samples $N$ phrases in $M$ intervals, with each interval containing $L$ phrases. $MN$ phrases is sampled in total. This process is shown in Figure~\ref{fig:retriever}. This method also guarantees the diversity of the phrase pool.

\subsubsection{Phrase Pool Augmentation}

To further augment the phrase pool, we apply additional strategies. The phrase pool is divided into upper and lower halves based on the source sentence position. If the source sentence of a phrase belongs to the upper half of the original lyrics paragraph, the phrase is also sorted into the upper half phrase pool, and vice versa. Similarly, when connecting phrases, the upper half phrase pool generates the upper half of the lyrics paragraph and vice versa. This strategy is based on the fact that ancient Chinese lyrics are usually divided into two paragraphs (Shang Que / Xia Que) with the same sentence structure. The upper half is more likely focused on describing the environment, while the lower half focuses on emotions.

\subsection{Phrase Connector}\label{PhraseConnector}

The phrase connector is an essential part of the system. It first searches for the designated song structure from a rhythmic-song structure look-up table, according to the provided rhythmic. The song structure is stored as a collection of tuples 
\begin{equation}
    r = \{(l_i, c_i)\}^N_{i=1}
\end{equation}
with $N$ specifies the number of sentences in the song. $l_i$ and $c_i$ specifies the number of words in sentence $i$ and the punctuation after sentence $i$, respectively. According to the song structure, the phrase connector selects a series of phrases from the constructed phrase pool that minimizes a multi-term loss function. 
The multi-term loss function consists of fluency loss, rhyme loss, and song structure loss. In section \ref{NextPhrase}, we show how the fluency loss is computed. In section \ref{Beam}, we illustrate the other losses and the minimizing algorithm.

\subsubsection{Next Phrase Prediction}\label{NextPhrase}

When connecting phrases, we want to determine whether $p_{im}$ and $p_{jn}$ can be connected fluently. This task is also worth additional discussions. First, while connecting two phrases, we might also have to consider the history context, maybe the entire generated history. Second, two phrases can be connected in different ways. Apart from a simple connection, two phrases might be connected with a comma inserted between or even a period between. If we eliminate these possibilities, it is unjustifiable to say that $p_{im}$ and $p_{jn}$ have no relationship if they cannot be connected directly. To tackle this issue, we adopt the BERT model\cite{bert_devlin_chang_lee_toutanova} and fine-tune the model on a novel downstream task called Next Phrase Prediction (NPP). This task is inspired by the Next Sentence Prediction (NSP) task used in BERT pretraining. Two sentences are sent into the BERT model during the pretraining stage, separated with a [SEP] token. The NSP task judges whether the two sentences could be connected. The NPP task follows a similar design. We construct the NPP input by concatenating the whole generated history context and the next candidate phrase with a [SEP] token. The training set is also constructed from the original ancient Chinese lyrics and phrase datasets. We collect training, validation, and testing samples by connecting two consecutive phrases $p_im$ and $p_jn$. Consecutive means the sampled phrases pair are either connected directly, with a comma, or with a period. For 25\% of the samples, we replace $p_{jn}$ with a random phrase. For the other three cases, samples are distributed uniformly. The training set can be denoted as
\begin{equation}
    \{(p_{im}, p_{jn}, Q(p_{im}, p_{jn}))\}
\end{equation}
where the group truth label $Q(p_{im}, p_{jn})$ is given by a one-hot vector:

\begin{equation}
    Q(p_{im}, p_{jn}) = 
    \left[
    \begin{array}{cc}
       \mathbbm{1}_{i=j} \cdot \mathbbm{1}_{n-m=1}    \\
          \mathbbm{1}_{j-i=1} \cdot \mathbbm{1}_{m=-1} \cdot \mathbbm{1}_{n=0} \cdot \mathbbm{1}_{i,j} \\
          \mathbbm{1}_{j-i=1} \cdot \mathbbm{1}_{m=-1} \cdot \mathbbm{1}_{n=0} \cdot \mathbbm{1}_{i.j} \\
          \mathbbm{1}_\text{otherwise}
    \end{array}
    \right]
\end{equation}

\begin{figure}[H]
	\begin{center}
		\includegraphics[scale=0.16]{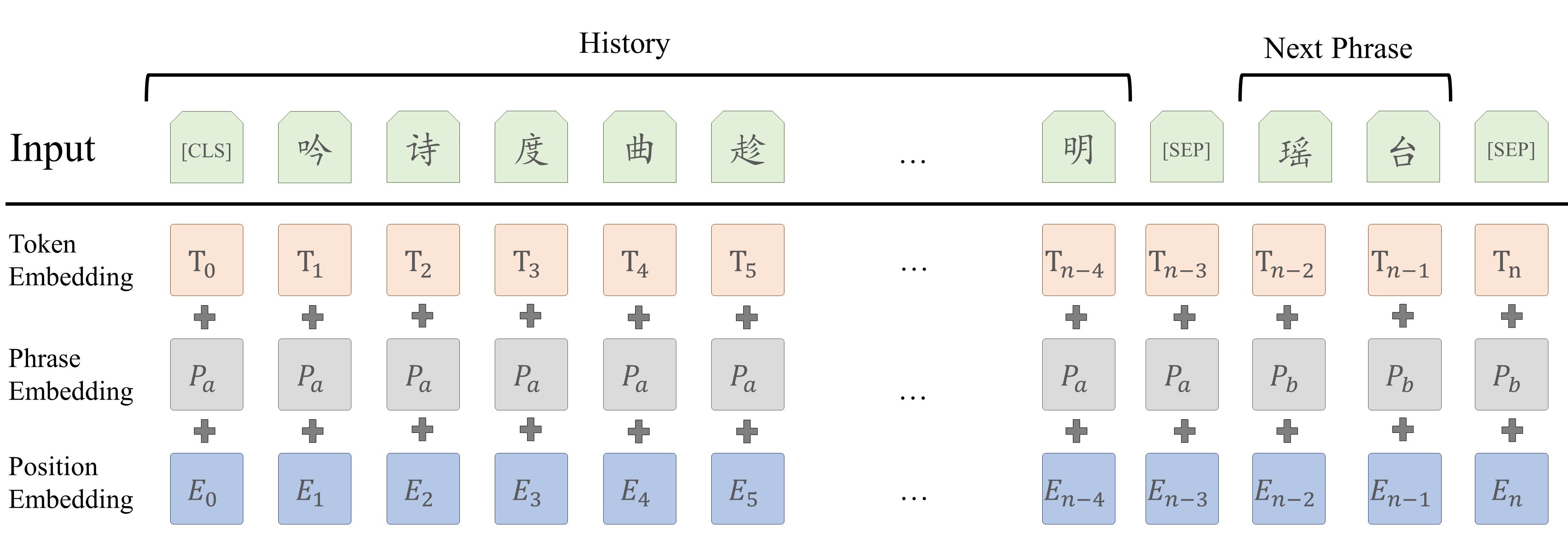}
	\end{center}
	\caption{Input of the BERT model for the NPP task}
	\label{fig:bertinput}
\end{figure}

$\mathbbm{1}_{i,j}$ and $\mathbbm{1}_{i.j}$ indicates sentence $s_i$ and $s_j$ are connected with a comma and period, respectively. Index $-1$ denotes the last phrase index in a sentence. Figure~\ref{fig:bertinput} shows an input sample of the NPP task.

We adopt the cross-entropy loss as the training objective and train the model $M$ with parameter $\theta$.

\begin{equation}
    \mathcal{L}(\theta) = \sum \text{M}(\theta, \textbf{p}, \textbf{p'}) \log Q(\textbf{p}, \textbf{p'})
\end{equation}

Finally, in the inference stage, given the history and a candidate phrase $p$, we define fluency loss $L^{flu}$ as

\begin{equation}
    \textbf{L}^\text{flu} = \text{M}(\text{history}, p)
\end{equation}

\subsubsection{Beam Search}\label{Beam}
We use the Beam Search algorithm to select a series of phrases that minimizes the fluency loss, the song structure loss, and the rhyme loss. The song structure loss expects the generated lyrics to have the same character number as the song structure in each sentence. For a given history and a candidate phrase, we calculate the song structure loss in all three scenarios: 1) the candidate phrase is connected directly; 2) the candidate phrase is connected with a comma inserted; 3) the candidate phrase is connected with a period inserted. The song structure loss $L^{\text{ss}}$ is given by

\begin{equation}
    L^{\text{ss}} = \sum \text{F}(\frac{|l_i-h_i|}{l_i}) 
\end{equation}
\begin{equation}
     \text{F}(x) = \left\{
	\begin{aligned}
	2x-x^2 \quad x<1\\
	1 \quad x>1\\
	\end{aligned}
	\right.
\end{equation}
where $l_i$ and $h_i$ denote the expected length and the actual length of sentence $i$. We choose a quadratic form to encourage the model to prefer an exact match.

We use rhyme loss $L^{\text{rhy}}$ to ensure the rhyme is consistent for each sentence (sentences that end with a period). We define a rhyme mapping function that maps each rhyme into a one-hot row vector. When calculating, all rhyming words are collected and mapped to form a matrix, with each row representing a rhyming word and its rhyme. We compute the variance of each column and then sum them together. This value is further divided by the max variance possible to re-scale the range of rhyme loss to [0,1]. Again, given the history and the phrase candidate, we compute the rhyme loss in all three scenarios.

Finally, we integrate all three losses and use Beam Search to select a series of phrases. 

\begin{equation}
    L = \alpha L^\text{flu}(\text{history}, p) + \beta L^{\text{ss}}(\text{history}, p) + \gamma L^{\text{rhy}}(\text{history}, p)
\end{equation}

To initialize the search process, we select all the phrases that appear at the beginning of its lyrics source to form the initial beams (history). In a single iteration, for each candidate and each history, we calculate the integrated loss for all three scenarios. However, once a phrase candidate is already used in the history, it will not be used again. This also ensures phrase diversity in the generated work. At the end of each iteration, the (history, phrase, connect method) pair with large loss will be pruned. Otherwise, the history and the phrase will be connected using the given method to form the new history and begin the next iteration.

\section{Results and Discussion}

We test our model on a prompt set and conduct three comparative experiments to validate our lyrics-generating system, one for intra-sentence fluency, one for sentence transition fluency, and one for the generation quality for the entire piece. The prompt set contains 100 topic and rhythmic pairs. The topics are single sentences randomly sampled from the Complete Ancient Chinese Lyrics dataset (Quan Song Ci) and the Complete Ancient Chinese Poems dataset (Quan Tang Shi). The rhythmics are sampled from a manually selected rhythmic set that contains ten famous rhythmics. In the following sections, we first show the baseline model is designed in section \ref{baseline}. Then
we present the results for the upper experiments in section \ref{Intra}, \ref{SentenceTransition}, and \ref{FullPiece}, respectively.

\subsection{Baseline Method}\label{baseline}

The baseline model for this specific task is hard to find. As mentioned in section \ref{AncientLyricsGeneration}, many insightful methods are focused on poems that do not require precise word number control in each sentence. Hence, we simply apply a naive approach as the baseline method. The naive method includes only one module, the sentence retrieving module, that would retrieve a series of sentences that are related to the input topic. The retrieving strategy is identical to the proposed method introduced in section \ref{PhraseRetriever}. The only difference is that instead of retrieving phrases, we now split the original paragraph into sentences and retrieve the sentences directly. After retrieval, the naive approach randomly select from the sentences to fit in the song structure (e.g., for a six-word sentence, the naive approach randomly selects from all six-word sentence retrieved). This method does not consider the sentence transition fluency and the rhyme consistency amount sentences. It also has no creativity in modifying the content within each sentence.
\subsection{Intra Sentence Fluency}\label{Intra}

Because our model generates sentences phrase-by-phrase, we conduct a survey to evaluate the fluency within a single sentence. Our survey has six entries of generated sentence results and six entries of human-written sentences. The generated and human-written sentences are randomly selected from all the generated paragraphs and the Ancient Chinese Lyrics dataset, respectively. Each subject is required to rate three entries of the generated sentence and three entries of the human-written sentence (chosen randomly) based on a five-point scale from 1 (very poor) to 5 (very high). The rating criterion, fluency, is described as how logical a sentence is. This evaluation does not involve the baseline method because the baseline does not modify the content in each sentence and achieves the same fluency level as human work. We can regard the fluency level for human-written sentences as the fluency level for the baseline.

\begin{figure}
	\begin{center}
		\includegraphics[scale=0.15]{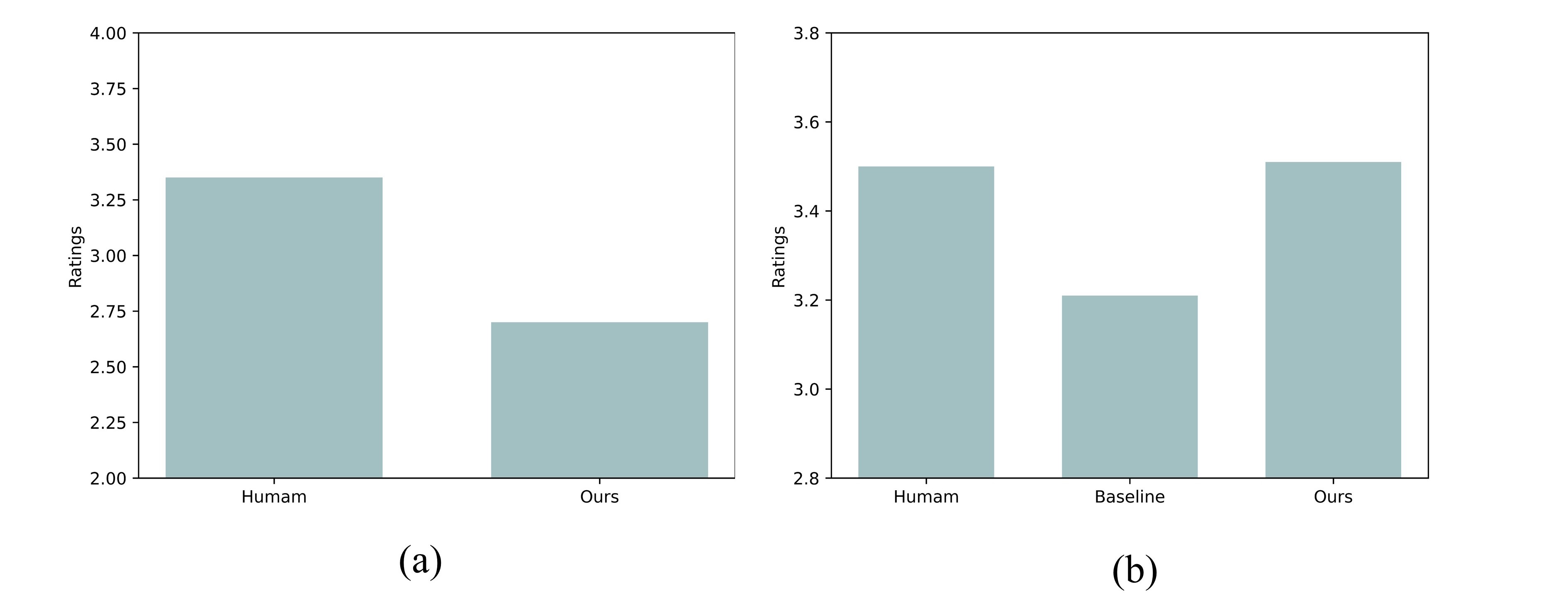}
	\end{center}
	\caption{The subject evaluation result of the experiments described in section \ref{Intra} and \ref{SentenceTransition}.}
	\label{graph:eva1}
\end{figure}

A total of 41 subjects with diverse literary backgrounds
participated in our survey, and all participants submitted an effective rating. As shown in Figure \ref{graph:eva1} (a), the height of the bars denotes the mean values of the ratings. We see that the logicality within the generated sentences is still significantly lower than in human-written sentences.

\subsection{Sentence Transition Fluency}\label{SentenceTransition}

We also conduct a survey to evaluate how fluent is the transition between sentences. The fluency also directly reflects the next phrase prediction power of the BERT model discussed in section \ref{NextPhrase}. This survey contains six human-written sentence pairs, six our model-generated sentence pairs, and six baseline-generated sentence pairs. The subject will rate on six entries where 2 of them are sampled from human-written paragraphs, two are sampled from our model's generation, and two are sampled from baseline's generation. Each subject will rate the sentence pairs using the five-point scale. The fluency criterion is described as how logical is the transition between the sentence pair. 
\begin{figure}
	\begin{center}
		\includegraphics[scale=0.06]{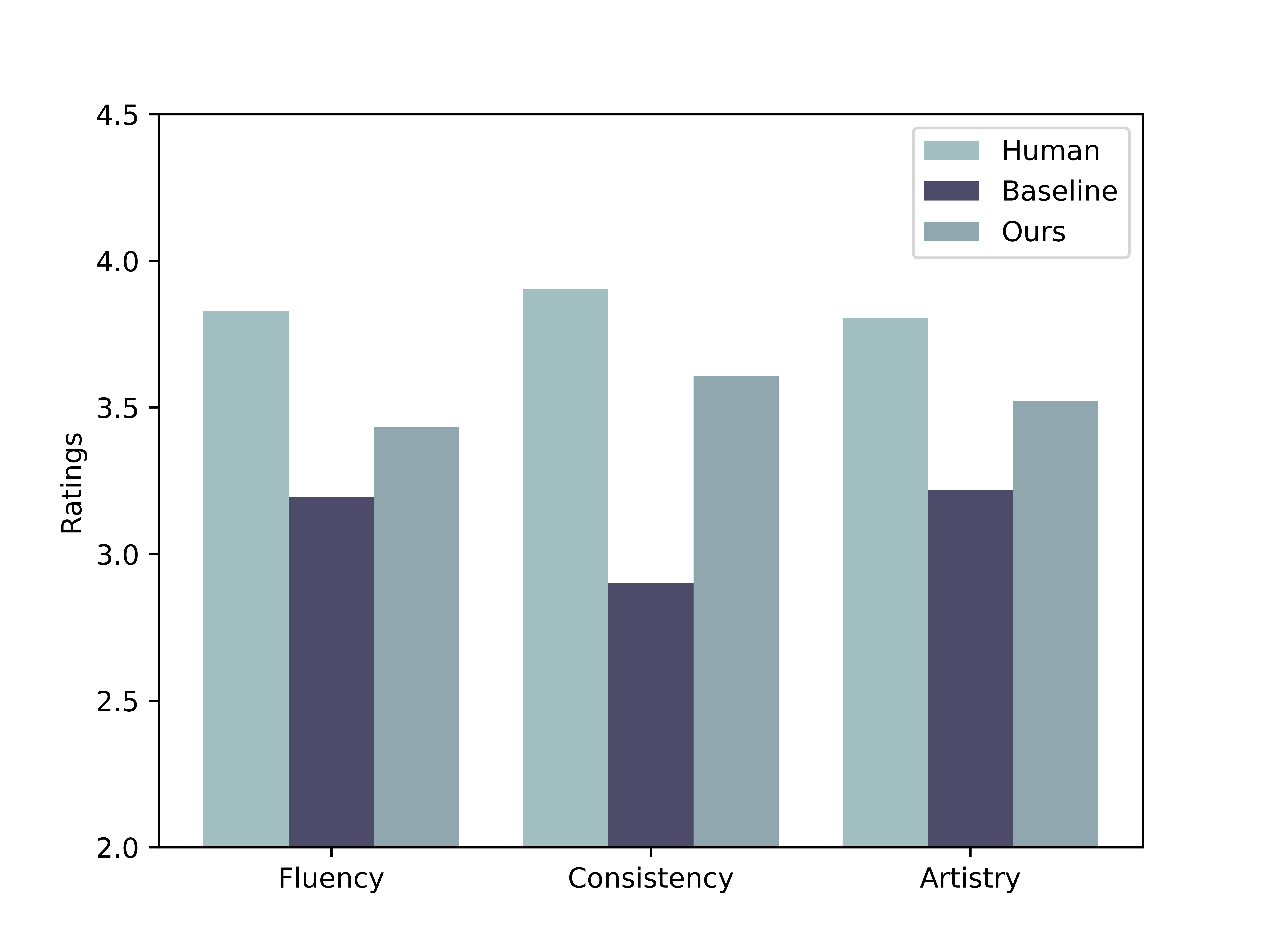}
	\end{center}
	\caption{The subject evaluation result of the experiments described in section \ref{FullPiece}.}
	\label{graph:eva2}
\end{figure}

Same with the last experiment, we collected 41 effective ratings, and the result is shown in Figure \ref{graph:eva1} (b). We see that our model achieves the same performance as the human poets. Both our model and human poets outperform the baseline significantly. The result is not surprising because the baseline method does not consider the transition between sentences.

\subsection{Full-piece Generation Result}\label{FullPiece}
Finally, we conduct another survey to evaluate the full-piece generation result. In our survey, each subject will see two human-written paragraphs (randomly chosen from 6), two paragraphs generated using the baseline method (randomly chosen from 4), and two paragraphs generated by our model (randomly chosen from 4). Apart from the lyrics, the subject will be provided with the prompts (topic and rhythmic) used by the generated results. For the human-written paragraph, we choose the first sentence as the topic because no real topic is available. The subject is required to rate the paragraphs based on the five-point scale and the following three criteria:
\begin{enumerate}
    \item Fluency: Whether the whole paragraph is logical and meaningful.
    \item Consistency: Whether the paragraph's content can fit the topic well.
    \item Artistry: How artistic is the generated result, including whether rhymes are consistent, how beautiful the lyrics are, etc.
\end{enumerate}

Same with the previous two experiments, a total of 41 effective ratings were collected, and the result is shown in Figure \ref{graph:eva2}. We see that our model outperforms the baseline in all three criteria. For fluency, human-written paragraphs receive the highest rating, and our model is marginally better than the baseline. This is probably because the baseline method achieves a high intra-sentence fluency but cannot write logical sentence transitions due to the nature of random assignment. As for consistency, our model outperforms the baseline significantly, and there is only a small gap between our model and human poets. However, the consistency of human-written paragraphs might be underestimated because we assume the first sentence is the topic. It is possible that the first sentence cannot reflect the general topic well. For Artistry, our model outperforms the baseline marginally, but the ratings for all three methods are close to each other.

\section{Discussion}\label{discussion}
According to the experimental results, our model achieves good generating power while guaranteeing controllability on song structure and topic. This approach can be seen as an auto-regressive method but using phrases as single tokens. The original motivation is the fact that lyrics are usually created by piling up literary images. We want to use phrase-level tokens to preserve those images in our vocabulary to achieve a better generating result and controlling power. In the actual implementation, the original GPT architecture is unsuitable because the nature of phrase-level tokens leads to a huge vocabulary size. Hence, we design a phrase retriever module that first constructs a small phrase pool from the general phrase database, then utilizes BERT and Beam Search to select a series of phrases that can connect to each other fluently while fitting in the song structure and consistent in rhyme. Compared with previous work, this approach archives high controllability in song structure control. Moreover, the current model is very extensive. More constraints can be added to the phrase pool to achieve complex topic control in sentence granularity.

However, some limitations also exist. First, the current model would take a long time to generate. A typical generation time cost is 1 minute (test on NYU HPC (Greene)). The phrase connecting process is responsible for this time cost. It searches phrases with a $O(n^2)$ complexity, and large-scale language model inference is required in each iteration. In some usage scenarios, this time cost might be unacceptable.

In addition, one major concern about this design is making the mismatch in song structure and rhyme a penalty instead of forcing the model to generate based on a fixed song structure and some rhymes. The current approach will result in some generated work not fitting in the song structure or not consistent in rhyme (e.g., if connecting two phrases has a low fluency loss, they might be connected even if the rhyme is not consistent or the number of characters is wrong). In some lyrics generating task, this is not a big problem, as the song structure and rhyme is not a compulsive constraint. However, for ancient Chinese lyrics, failing to meet the constraint will result in the nullification of the whole work. This research sticks to the current design because there should be an additional rewriting step after phrase connecting in the original proposal. The sentence rewriting module would rewrite the generated sentences to guarantee that the number of words and rhymes are correct. Unfortunately, we do not have enough time to finish it due to the time limit.
\section{Conclusion}

This report proposes a novel method for generating ancient Chinese lyrics. The system is equipped with a phrase retriever that constructs a phrase pool from a self-collected phrase database according to an input prompt and a phrase connector that select a series of phrases from the phrase pool to optimize a multi-term loss function. Our main contributions include a novel phrase-retrieving-based method for lyrics generation and a phrase-connecting algorithm that can generate high-quality and controllable lyrics.

In the future, we plan to further optimize the model by 1) adding a sentence rewriting module to ensure rhyme consistency and song structure integrity; 2) speeding up the generation.

\newpage
\singlespacing
\bibliographystyle{IEEEtran}
\bibliography{references}


\end{document}